# Detector Algorithms of Bounding Box and Segmentation Mask of a Mask R-CNN Model

Haruhiro Fujita
*Faculty of Management and Information Sciences*
Niigata University of International and Information Studies
Niigata City, Japan
fujita@nuis.ac.jp

Masatoshi Itagaki
*BSN iNet Co. Ltd.*
Niigata City, Japan
masa-ita@bsnnet.co.jp

Yew Kwang Hooi
*Department of Computer and Information Sciences*
Bandar Sri Iskandar, Malaysia
yewkwanghooi@utp.edu.my

Kenta Ichikawa
*BSN iNet Co. Ltd.*
Niigata City, Japan
ichikawa@bsnnet.co.jp

Kazutaka Kawano
*Department of Curatorial Research*
Tokyo National Museum tudies
Tokyo, Japan
kawano4191kka489@gmail.com

Ryo Yamamoto
*Department of Cutorial Planning*
Tokyo National Museum
Tokyo, Japan
yamamoto-r62@nich.go.jp

*Abstract*—Detection performances on bounding box and segmentation mask outputs of Mask R-CNN models are evaluated. There are significant differences in detection performances of bounding boxes and segmentation masks, where the former is constantly superior to the latter. Harmonic values of precisions and recalls of linear cracks, joints, fillings, and shadows are significantly lower in segmentation masks than bounding boxes. Other classes showed similar harmonic values. Discussions are made on different performances of detection metrics of bounding boxes and segmentation masks focusing on detection algorithms of both detectors.

*Keywords— road object detection, Mask R-CNN models, mAP, AR, harmonic means, detection algorithms of detectors*

## I. Introduction

Object detection is a type of cognitive image processing under the field of computer vision [1]. An efficient object detection is a main challenge for autonomous driving in modern traffic systems which requires advanced deep learning models to learn the environment and to control the system [2]. Many models have been developed for object detection, but few may be suitable for object detection on a moving vehicle.

In recent years, better models have allowed for more faster object detection. Region Convolutional Neural Network (R-CNN) detects objects in images by drawing one or more bounding boxes. The model extracts multiple windows based on adjacent pixels' texture, colour or intensity in sliding window for CNN features computation prior to Support Vector Machine classification. The CNN computation is done using AlexNet.

On R-CNN, the CNN features computation is very costly resulting in slow training. For every image, determining region of interest may reach up to 2000 forward passes involving training of 3 separate models. Thus, Fast R-CNN was introduced to streamline the pipeline. The algorithm combines all models and limits the 2000 forward passes to the original and then share the computation for all overlapping sub-regions through ROI pooling technique. Through combined models, feature extractor, classifier and regressor are joint together in a unified framework. The region proposal created using selected search in Fast R-CNN is a bottle neck and hence an opportunity improved in Faster R-CNN. Faster R-CNN optimizes selective search algorithm by reusing features of the images that are already calculated in CNN for region proposals. Thus, both region proposal and classification are carried out simultaneously resulting in only one CNN training. The latest model is Mask R-CNN which provides pixel-level segmentation instead of bounding box. Mask R-CNN is developed by adding a full CNN (FCN) on top of CNN feature map of a Faster R-CNN. Pixel that belong inside the object is marked as 1 and every else as 0, i.e. mask. Various Mask R-CNN models have been introduced since 2017. There were some outstanding issues such as accuracy of region of interest due to misalignment and the higher computing cost.

This manuscript is organized as follows:- literature review of surface object detection technologies and applications for road surface object detections; specific objectives of this study; methods of study; result of the testing the models; discussion of the profiles of each model and pinpointing critical areas for improvement.

## II. Literature Review

This section describes the data selection and preparation for the study and assessing existing R-CNN APIs, models and studies in surface object detection particularly for road surface assessment.

### A. Mask Region Convolutional Neural Network (Mask R-CNN)

Instance segmentation detects an object under overlap of other objects precisely. However, it does not label all of the pixels of the image, as it segments only the region of interests [6]. The state-of-the-art instance segmentation approach is the detection-based method that predicts the mask for each region after acquiring the instance region [7].

[8] proposed Mask Region Convolutional Neural Network (hereafter Mask R-CNN). The Mask R-CNN is an extended model of the Faster R-CNN [9]. Faster R-CNN is a forerunner of instance segmentation but limited to detect object in bounding boxes only. On the other hand, Mask R-CNN can detect objects in segmentations.

Mask R-CNN adds learning segmentation masks (called Mask Branch) sub-module onto Faster R-CNN. The sub-module predicts segmentation masks on each Region of

Interest (RoI) [6] in parallel with each other using convolution arrays of data for classification and bounding box regression. The mask module is a small fully convolutional network (FCN) applied to each RoI, predicting a segmentation mask for each pixel.

*B. Road damage detection models*

The previous study [11] evaluated road surface object detection tasks using four Mask R-CNN models available on the TensorFlow Object Detection API. The models were pre-trained using COCO datasets and fine-tuned by 15,188 segmented road surface annotation tags. Test data set was used to obtain Average Precisions and Average Recalls. Result indicated a substantial false negatives or "left judgement" counts for linear cracks, joints, fillings, potholes, stains, shadows and patching with grid cracks classes. There were significant number of incorrectly predicted label instances.

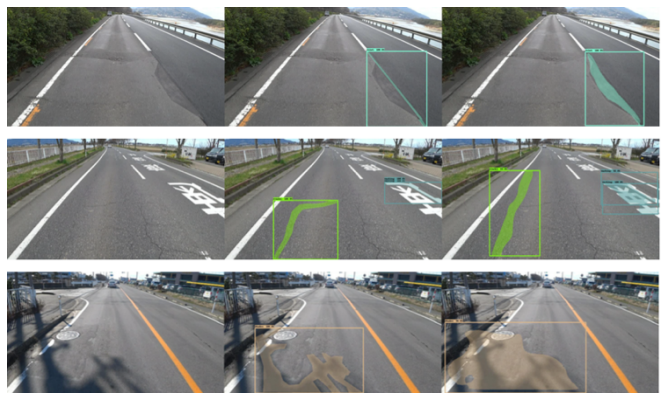

Fig. 1. Road surface object detection using one of Mask R-CNN models

### III. OBJECTIVES

The objectives of this study are to:-

- compare inferences of bounding boxes and segmentation masks
- generates harmonic means of precisions and recalls to evaluate performances
- develop discussion on algorithm architectural details

### IV. METHODOLOGY

Methodologies were the same as previously reported with the following brief descriptions [11].

*A. Data and Model Preparation*

Refer to TABLE I. The annotation data contains 12 classes of road damages and cognitive objects in the road images to differentiate those damages from others. Annotations in those 1,418 road color images with a size of 3840 x 2160 were total of 15,188. All visible objects of the predefined classes in two third bottom part of each image were segmented and tagged.

TABLE I. ROAD OBJECT ANNOTATION DATA OBTAINED FROM 1418 ROAD IMAGES FOR MODEL FINETUNING

| classes | tag | Datasets/segmentations | | | |
| --- | --- | --- | --- | --- | --- |
| | | training | validation | testing | total |
| Linear Cracks | Crack1 | 2,315 | 468 | 455 | 3,238 |
| Grid Cracks | Crack2 | 524 | 103 | 101 | 728 |
| Pavement Joints | Joint | 941 | 198 | 219 | 1,358 |
| Patchings | Patching | 377 | 102 | 77 | 556 |
| Fillings | Filling | 1,240 | 207 | 187 | 1,634 |
| Pot-holes | Pothole | 143 | 39 | 14 | 196 |
| Manholes | Manhole | 296 | 61 | 52 | 409 |
| Stains | Stain | 98 | 22 | 12 | 132 |
| Shadow | Shadow | 1,084 | 252 | 212 | 1,548 |
| Pavement Markings | Marking | 1,239 | 241 | 297 | 1,777 |
| Scratches on Markings | Scratch | 2,433 | 494 | 576 | 3,503 |
| Grid Crack in Patchings | Patching2 | 73 | 19 | 17 | 109 |
| Total | | 10,763 | 2,206 | 2,219 | 15,188 |

*B. Computation Resources*

The computational analyses were conducted with a machine of Intel Core i7-8700K(6cores/12 threads/3.70GHz) CPU, 32GB CPU memory, Nvidia GeForce GTX 1080 GPU with 2560 CUDA core, GPU clocks of 1607/1733 MHz, 8GB GDDRSX GPU memory and 320GB/s memory bandwidth. The language for programming is Python 3.6.9 using libraries from Object Detection API version 1 on top of TensorFlow 1.15.0.

*C. Methods*

Main steps of the methods are performing annotation, mask data making; model training and fine tuning.

*1) Annotation and Mask Data Making:*

Refer to Fig. 2. The files output by the Visual Object Tagging Tool (VoTT) were converted to intermediate annotation files. Next, the files were converted into TensorFlow Record files for training, validation and testing. The conversions were done using three separate conversion script executive programs in Python. The last conversion script (vott2tfrecords.py) modified the original image size of 3840x2160 (width x height) to match with the training image size of 960x540 used in machine learning.

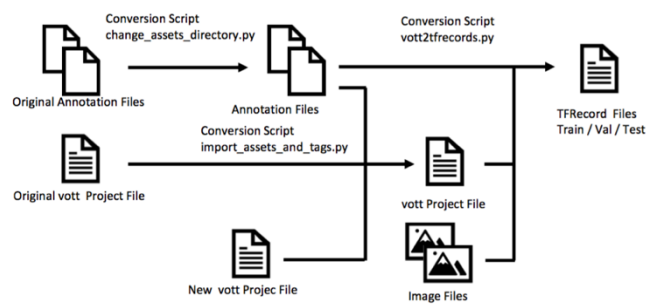

Fig. 2. Flow for annotation and data analysis.

*2) Model Training and Fine Tuning:*

Refer to Fig. 3. Pre-trained model checkpoint file, TensorFlow Record files, ground truth annotation data/labels ("tf_label_map.pbtxt") and pipeline.config are used for training of object detection. The pretrained model checkpoint is based on MS COCO dataset. The TensorFlow Record files contain both datasets for training and validation. The training generated Fine Tuned Checkpoint file from which an object detection model was acquired. Finally, the fine tuned detection model was evaluated using validation and testing datasets. The inference result produces precision metric, recall metric and confusion matrix. The training was done for each four Mask R-CNN models.

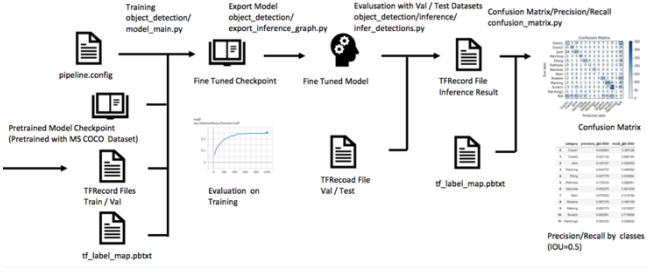

Fig. 3. Training and evaluation processes of pre-trained Mask R-CNN models.

*D. Model Evaluation Metrics*

The current work in progress is a preliminary analysis. Thus, the model evaluation result was limited to validation datasets. Later works will use test datasets.

Various metrics used for model evaluation of this work are mean average precisions (mAP) and average recalls (AR) at various levels of Intersection on Unit(IoU). The Mask R-CNN has a region proposal network layer that makes multiple inferences simultaneously on the class classification, the segmentation and the mask areas resulting six loss metrics.

Besides above model-wise metrics, the average precisions and the average recalls at IoU=0.5 are used for all twelve road object classes.

## V. RESULTS

*A. Comparison of overall metrics for bounding box and segmentation masks*

For easier model identification, they are named as follows:

- Model #1: Mask R-CNN Inception v2 coco;
- Model #2: Mask R-CNN Inception ResNet v2 Atrous coco;
- Model #3: Mask R-CNN Inception ResNet50 Atrous coco;
- Model #4: Mask R-CNN Inception ResNet101 Atrous coco.

Refer Table II. It shows the AP(precision) and AR(recall) metrics in bounding boxes and instance segmentations respectively of those four models. The highest value in each metrics class is shown in bold letters. The best mAP, $AP_{50}$ and $AP_{75}$ in bounding boxes were achieved by the model #3, 0.2810, 0.4789 and 0.2965 respectively. The best AR@10 and AR@100 in bounding boxes were achieved by the model #4 of 0.4238 and 0.4621 respectively, though the best AR@1 was done by the model #2 of 0.2401. For the detection of instance segmentations, the best mAP, $AP_{75}$, AR@10 and AR@100 were achieved by the model #4, 0.2118, 0.2045, 0.3265 and 0.3438 respectively. Similarly the best AR@1 was achieved by the model #2 of 0.2002, as in bounding boxes. The significant higher all APs and ARs metrics values of bounding boxes than of instance segmentations is notable.

*B. Comparison of class metrics for bounding boxes and segmentation masks*

*1) Harmonic mean (F1)*

The definitions of F1 in object detection are as followed,

$$Precison = \frac{True\ Positives}{True\ Positives\ +\ False\ Positives}$$

$$Recall = \frac{True\ Positives}{True\ Positives\ +\ False\ Negatives}$$

$$F1 = \frac{True\ Positives}{True\ Positives + \frac{1}{2}(False\ Positives + False\ Negatives)}$$

The F1 was calculated as an integrated metrics of precision and recall, and the results of class detections are shown in Table III and Table IV for bounding boxes and instance segmentations respectively, with bold letters for the class highest values.

*2) Bounding box detection*

For the bounding box detection (Table III), model #4 was superior in AP@IoU=0.5, with the highest precisions in five classes, followed by model #3 in four classes, then model #1. Model #3 was superior in AR@IoU=0.5, with highest recalls in five classes, followed by model #2 and model #4 in three classes each. Model #2 showed the highest harmonic means (F1) in five classes followed by model #3 in four classes.

*3) Segmentation mask detection*

For the segmentation mask detection (Table IV), model #2 was superior in AP@IoU=0.5, with the highest precisions in five classes, followed by model #1 in three classes. Model #2 was superior in AR@IoU=0.5, with highest recalls in seven classes, followed by model #4 in four classes. Model #2 showed the highest harmonic means (F1) in six classes followed by model #3 and model #4 in three classes each.

TABLE II. MODEL METRICS FOR BOUNDING BOXES AND INSTANCE SEGMENTATIONS

| Models | Bounding Box | | | Segmentation Mask | | | Bounding Box | | | Segmentation Mask | | |
|---|---|---|---|---|---|---|---|---|---|---|---|---|
| | AP | $AP_{50}$ | $AP_{75}$ | AP | $AP_{50}$ | $AP_{75}$ | AR@1 | AR@10 | AR@100 | AR@1 | AR@10 | AR@100 |
| Mask R-CNN Inception v2 coco | 0.2432 | 0.4382 | 0.2482 | 0.1600 | 0.3257 | 0.1279 | 0.1995 | 0.3765 | 0.4140 | 0.1559 | 0.2533 | 0.2713 |
| Mask R-CNN Inception ResNet v2 Atrous coco | 0.2756 | 0.4678 | 0.2815 | 0.2096 | 0.4038 | 0.1982 | **0.2401** | 0.4215 | 0.4368 | **0.2001** | 0.3212 | 0.3296 |
| Mask R-CNN ResNet50 Atrous coco | **0.2810** | **0.4789** | **0.2965** | 0.2104 | 0.4140 | 0.1934 | 0.2282 | 0.4147 | 0.4599 | 0.1895 | 0.3172 | 0.3417 |
| Mask R-CNN ResNet101 Atrous coco | 0.2779 | 0.4787 | 0.2853 | **0.2118** | 0.4036 | **0.2045** | 0.2311 | **0.4238** | **0.4621** | 0.1936 | **0.3265** | **0.3484** |

TABLE III. AP(IoU=0.5), AR(IoU=0.5) AND F1 OF 12 OBJECT CLASSES FOR BOUNDING BOX

Bounding Box

| Precision | Crack1 | Crack2 | Joint | Patching | Filling | Pothole | Manhole | Stain | Shadow | Marking | Scratch | Paching2 |
|---|---|---|---|---|---|---|---|---|---|---|---|---|
| Mask R-CNN Inception v2 coco | **0.4085** | 0.4958 | 0.3602 | **0.6346** | 0.4773 | **0.5714** | 0.8298 | 0.0000 | 0.3975 | 0.6224 | 0.5934 | 0.4000 |
| Mask R-CNN Inception ResNet v2 Atrous coco | 0.3791 | **0.5780** | 0.4328 | 0.5781 | 0.5000 | 0.5455 | 0.9167 | 0.0909 | 0.4870 | 0.6268 | 0.6391 | 0.3000 |
| Mask R-CNN ResNet50 Atrous coco | 0.3050 | 0.4024 | 0.4286 | 0.6066 | 0.4229 | 0.5385 | **0.9756** | **0.2222** | 0.4424 | 0.6232 | **0.6509** | **0.5000** |
| Mask R-CNN ResNet101 Atrous coco | 0.3861 | 0.4962 | **0.4624** | 0.6154 | **0.5195** | 0.5000 | 0.9111 | 0.1538 | 0.4889 | **0.6440** | 0.6162 | **0.5000** |

| Recall | Crack1 | Crack2 | Joint | Patching | Filling | Pothole | Manhole | Stain | Shadow | Marking | Scratch | Paching2 |
|---|---|---|---|---|---|---|---|---|---|---|---|---|
| Mask R-CNN Inception v2 coco | 0.2549 | 0.5842 | 0.3881 | 0.4286 | 0.3369 | 0.2857 | 0.7500 | 0.0000 | 0.3019 | **0.6162** | 0.6233 | 0.1176 |
| Mask R-CNN Inception ResNet v2 Atrous coco | **0.3275** | 0.6238 | **0.3973** | 0.4805 | 0.4171 | 0.4286 | **0.8462** | 0.0833 | 0.3538 | 0.5825 | 0.6701 | 0.1765 |
| Mask R-CNN ResNet50 Atrous coco | 0.3077 | **0.6733** | 0.3836 | 0.4805 | 0.4545 | 0.5000 | 0.7692 | **0.1667** | 0.4528 | 0.5960 | 0.6701 | **0.2941** |
| Mask R-CNN ResNet101 Atrous coco | 0.3055 | 0.6535 | 0.3653 | **0.5195** | 0.4278 | 0.5000 | 0.7885 | **0.1667** | 0.4151 | 0.5421 | **0.6858** | 0.2353 |

| F1 | Crack1 | Crack2 | Joint | Patching | Filling | Pothole | Manhole | Stain | Shadow | Marking | Scratch | Paching2 |
|---|---|---|---|---|---|---|---|---|---|---|---|---|
| Mask R-CNN Inception v2 coco | 0.4601 | 0.5545 | 0.4000 | 0.5426 | 0.4702 | 0.3810 | 0.8081 |  | 0.3968 | **0.6430** | 0.6520 | 0.2727 |
| Mask R-CNN Inception ResNet v2 Atrous coco | **0.5448** | 0.6095 | 0.4571 | 0.5532 | **0.5714** | 0.4800 | **0.9000** | 0.1739 | 0.4536 | 0.6213 | **0.7085** | 0.2222 |
| Mask R-CNN ResNet50 Atrous coco | 0.4923 | 0.5111 | **0.4723** | 0.5507 | 0.5309 | **0.5185** | 0.8602 | **0.1905** | 0.4569 | 0.6299 | 0.6963 | **0.3704** |
| Mask R-CNN ResNet101 Atrous coco | 0.4957 | 0.5726 | 0.4643 | **0.5775** | 0.5689 | 0.5000 | 0.8454 | 0.1600 | **0.4898** | 0.6325 | 0.6869 | 0.3200 |

TABLE IV. AP(IoU=0.5), AR(IoU=0.5) AND F1 OF 12 OBJECT CLASSES FOR INSTANCE SEGMENTATION

Segmentation Mask

| Precision | Crack1 | Crack2 | Joint | Patching | Filling | Pothole | Manhole | Stain | Shadow | Marking | Scratch | Paching2 |
|---|---|---|---|---|---|---|---|---|---|---|---|---|
| Mask R-CNN Inception v2 coco | 0.2285 | 0.4622 | 0.1200 | **0.6863** | 0.1742 | **0.5714** | 0.8298 | 0.0000 | 0.1987 | 0.3767 | 0.3565 | **0.6000** |
| Mask R-CNN Inception ResNet v2 Atrous coco | 0.3571 | **0.5321** | **0.2814** | 0.5938 | **0.3910** | 0.5455 | 0.9167 | 0.0909 | **0.4133** | 0.5942 | **0.5937** | 0.2000 |
| Mask R-CNN ResNet50 Atrous coco | 0.2549 | 0.3846 | 0.2653 | 0.6393 | 0.2687 | 0.5385 | **0.9756** | **0.2222** | 0.2811 | 0.5493 | 0.5160 | 0.5000 |
| Mask R-CNN ResNet101 Atrous coco | **0.3593** | 0.5038 | 0.2733 | 0.6308 | 0.3182 | 0.5000 | 0.9111 | 0.1538 | 0.3143 | **0.6400** | 0.5523 | 0.5000 |

| Recall | Crack1 | Crack2 | Joint | Patching | Filling | Pothole | Manhole | Stain | Shadow | Marking | Scratch | Paching2 |
|---|---|---|---|---|---|---|---|---|---|---|---|---|
| Mask R-CNN Inception v2 coco | 0.1341 | 0.5446 | 0.1233 | 0.4545 | 0.1230 | 0.2857 | 0.7500 | 0.0000 | 0.1415 | 0.3704 | 0.3646 | 0.1765 |
| Mask R-CNN Inception ResNet v2 Atrous coco | **0.3077** | 0.5743 | **0.2557** | 0.4935 | **0.3262** | 0.4286 | **0.8462** | 0.0833 | **0.2925** | 0.5522 | **0.6215** | 0.1176 |
| Mask R-CNN ResNet50 Atrous coco | 0.2571 | 0.6436 | 0.2374 | 0.5065 | 0.2888 | **0.5000** | 0.7692 | **0.1667** | 0.2877 | 0.5253 | 0.5313 | **0.2941** |
| Mask R-CNN ResNet101 Atrous coco | 0.2835 | **0.6634** | 0.2146 | **0.5325** | 0.2620 | **0.5000** | 0.7885 | **0.1667** | 0.2594 | 0.5387 | 0.6146 | 0.2353 |

| F1 | Crack1 | Crack2 | Joint | Patching | Filling | Pothole | Manhole | Stain | Shadow | Marking | Scratch | Paching2 |
|---|---|---|---|---|---|---|---|---|---|---|---|---|
| Mask R-CNN Inception v2 coco | 0.1690 | 0.5000 | 0.1216 | 0.5469 | 0.1442 | 0.3810 | 0.7879 | - | 0.1653 | 0.3735 | 0.3605 | 0.2727 |
| Mask R-CNN Inception ResNet v2 Atrous coco | **0.3306** | 0.5524 | **0.2679** | 0.5390 | **0.3557** | 0.4800 | **0.8800** | 0.0870 | **0.3425** | 0.5724 | **0.6073** | 0.1481 |
| Mask R-CNN ResNet50 Atrous coco | 0.2560 | 0.4815 | 0.2506 | 0.5652 | 0.2784 | **0.5185** | 0.8602 | **0.1905** | 0.2844 | 0.5370 | 0.5235 | **0.3704** |
| Mask R-CNN ResNet101 Atrous coco | 0.3170 | **0.5726** | 0.2404 | **0.5775** | 0.2874 | 0.5000 | 0.8454 | 0.1600 | 0.2842 | **0.5850** | 0.5818 | 0.3200 |

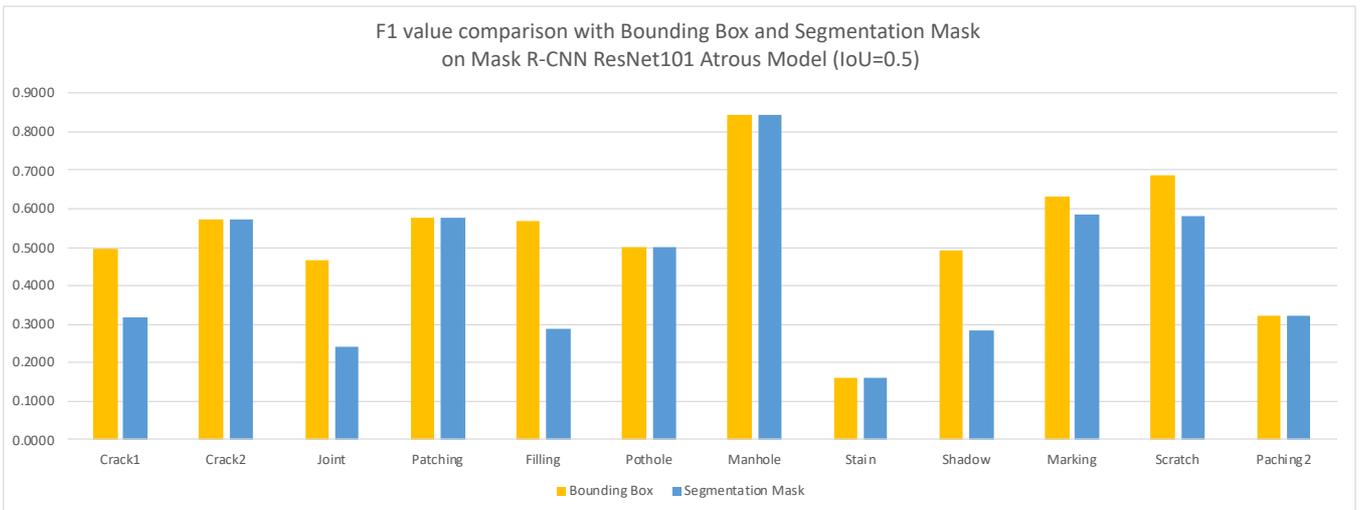

Fig. 4. Class F1 values of bounding boxes and segmentation masks of model #4

*4) Class F1 values of bounding boxes and segmentation masks*

Figure 4 shows comparison of class F1 values of bounding boxes and instance segmentations. F1 values of linear cracks, joints, fillings, and shadows are significantly lower in instance segmentations than bounding boxes. Other classes showed similar F1 values.

## VI. DISCUSSIONS

### A. Detection ability of the instance segmentation detector

The results showed the significant inferior F1 values of the mask segmentations than bounding boxes, in linear cracks, joints, fillings and shadows, which implies the model's less ability to learn instance segments. The common features of those four classes can be defined as elongated and the shapes of objects not fixed, as well as occupying wider areas. This less learning ability of the model #4 is also similar to other three models.

Although those Mask R-CNN models infer both rectangle bounding boxes and segmentation masks of bitmaps, the structure of the network is the same up to the part that selects RoI (Region of Interest), and after that, a branch that infers the position and size of the bounding boxes and the class of the object, and a branch that predicts the bitmap of segmentation masks, are connected.

As seen from the results of this test, the prediction performance of the segmentation masks tends to be lower than that of bounding boxes, due to the performance difference between those two predictors in the latter structure of the model, of which segmentation mask is more difficult task than bounding box task.

To demonstrate the above performance differences, the following configurations were set and those images with bounding boxes and segmentation masks were shown (figures 5 to 7).

・both confidence score and the IoU threshold were fixed at 0.5
・against ground truth annotations,
・IoU of bounding boxes are more than the threshold of 0.5
・IoU of segmentation masks are less than 0.5

### B. Algorithm architectural details of two detectors

The followings are all output files of the model including the outputs from the intermediate layers.

```
'rpn_box_predictor_features',
'rpn_features_to_crop',
'image_shape',
'rpn_box_encodings',
'rpn_objectness_predictions_with_background',
'anchors',
'refined_box_encodings',
'class_predictions_with_background',
'proposal_boxes',
'box_classifier_features',
'proposal_boxes_normalized',
'final_anchors',
'num_proposals',
'detection_boxes',
```

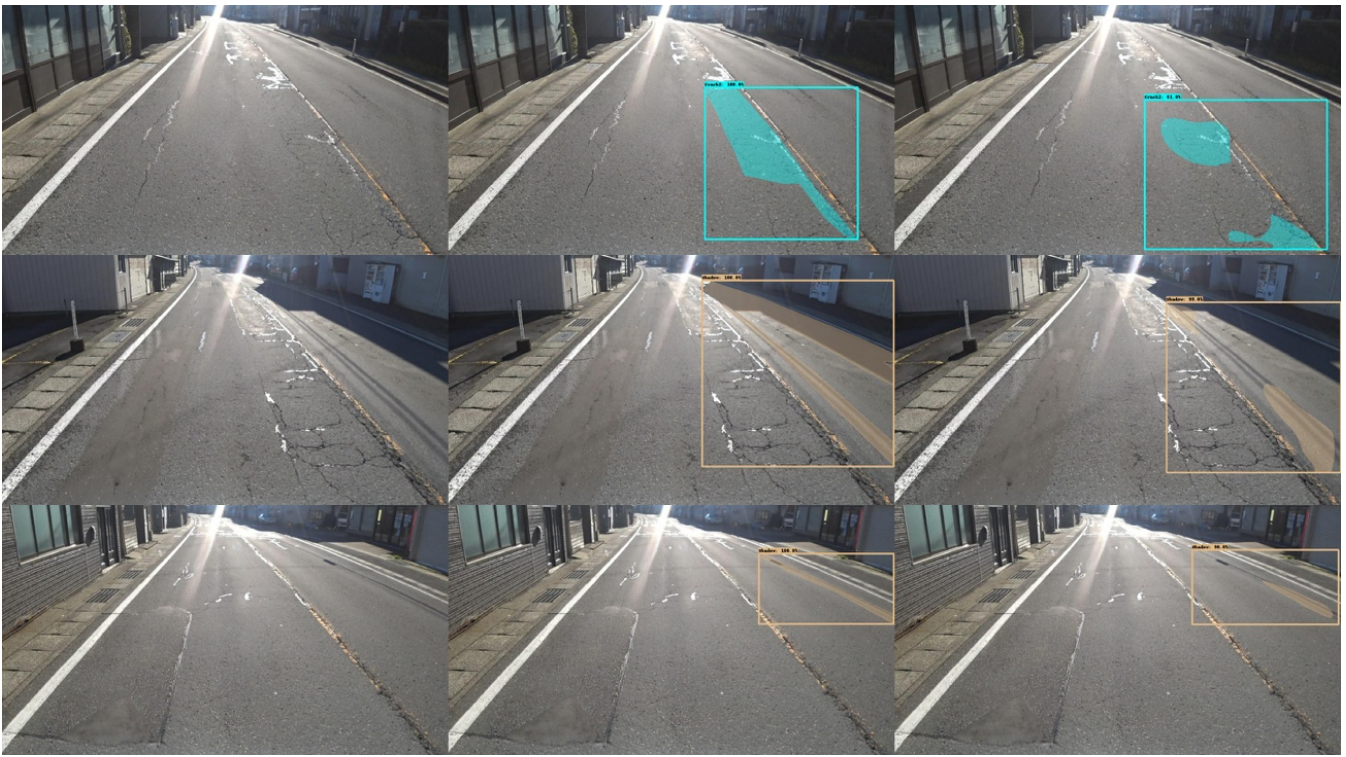

Fig. 5. Higher IoU of bounding boxes vs part of segmentation masks detected
(images left: original images, middle: boundary boxes, right segmentation masks)

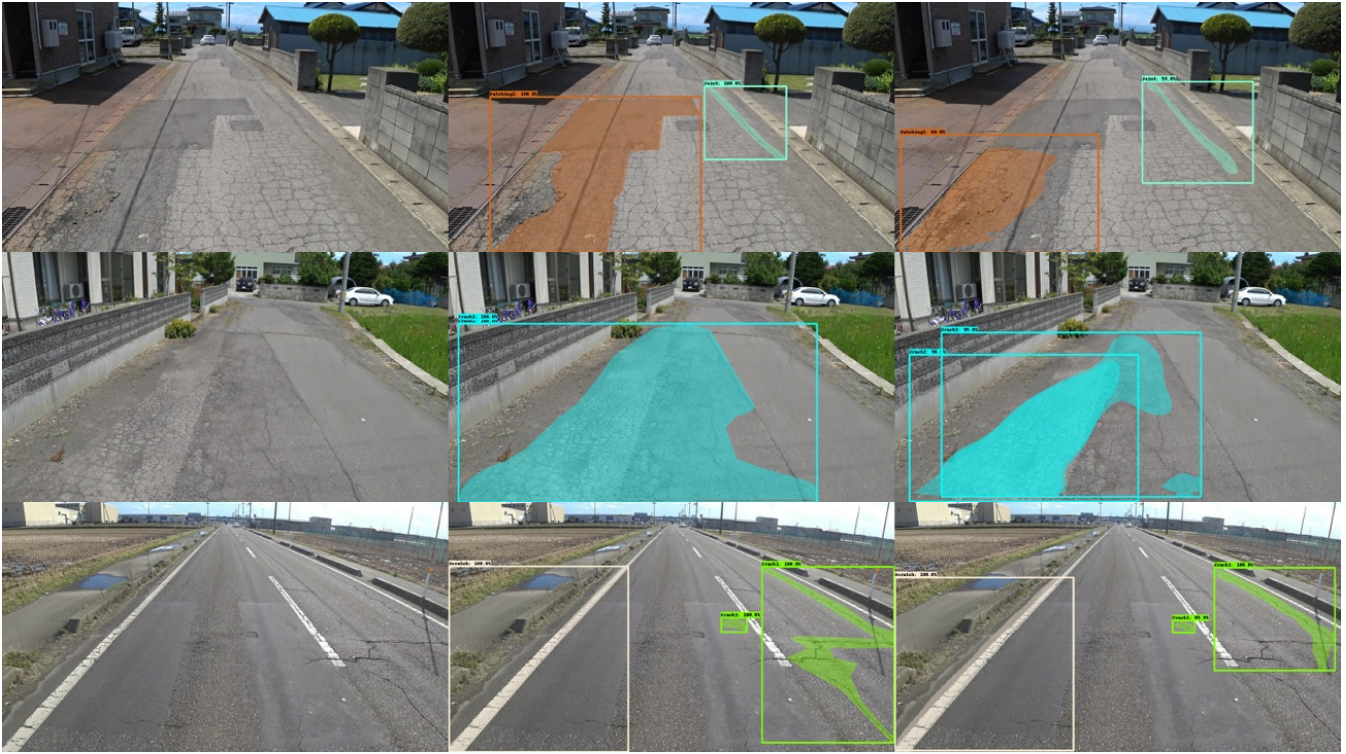

Fig. 6. IoU of bounding boxes > 0.5 vs IoU of segmentation masks <0.5
(images left: original images, middle: boundary boxes, right segmentation masks)

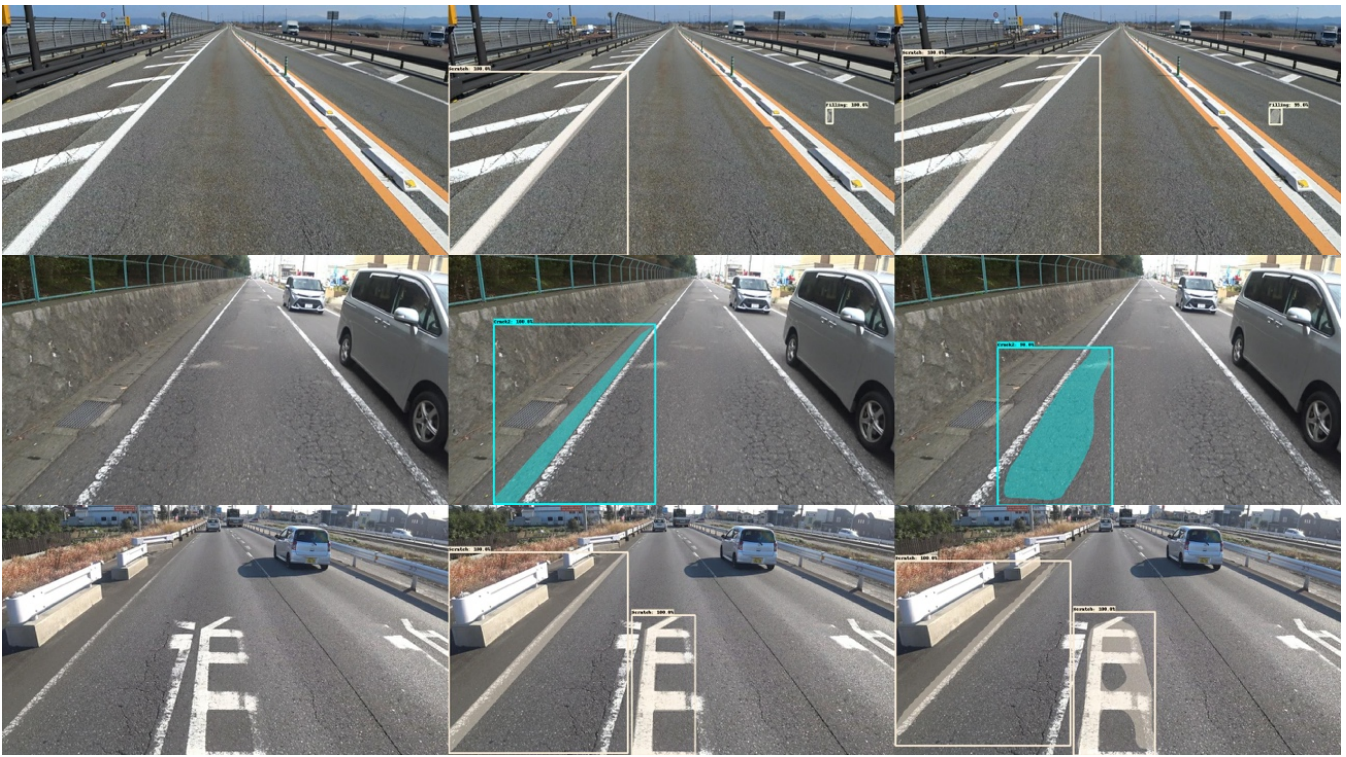

Fig. 7. Segmentation masks do not much
(images left: original images, middle: boundary boxes, right segmentation masks)

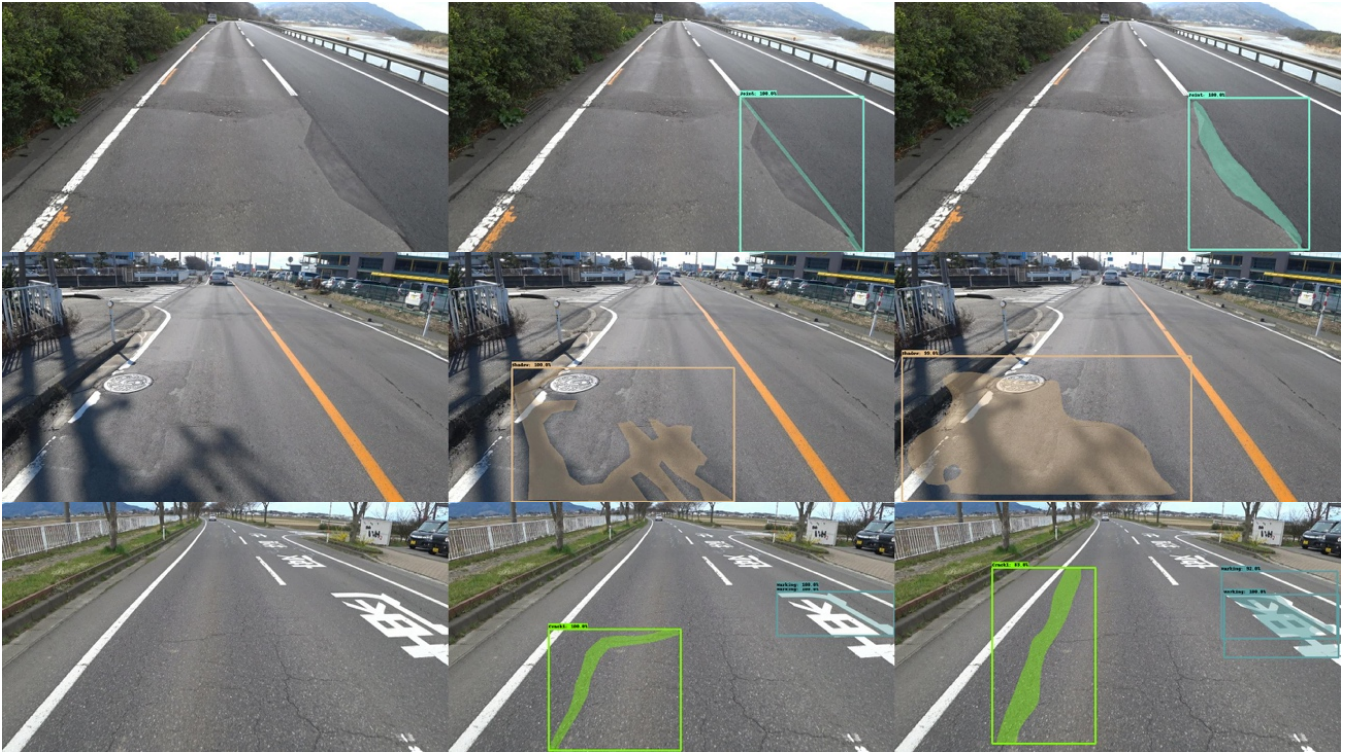

Fig. 8. Segmentation masks bigger than ground truths
(images left: original images, middle: boundary boxes, right segmentation masks)

'detection_scores',
'detection_classes',
'detection_multiclass_scores',
'detection_anchor_indices',
'num_detections',
'raw_detection_boxes',
'raw_detection_scores',
'mask_predictions',
'detection_masks'

The final outputs from the bounding box detector and the segmentation detector are as followed.

'detection_boxes',
'detection_scores',
'detection_classes',
'num_detections',
'detection_masks'

A schematic illustration of the algorithm mechanism outline is made as in the following figure (Fig.9).

*1) Bounding box detector*

The geometry of x and y coordinates of the center, height and width of bounding box is output (as in "X, Y, W, H"). The class definition was set to adding one more class as a background on top of the actual classes (13 classes altogether in this study, described as "B" in the figure). The sigmoid function output as "raw_detection_scores" is transfered to the next layer and converted as softmax function outputs in those 13 classes as "detection_multiclass_scores". Then the maximum probability class as "argmax detection class" and the certainty as "detection_scores" are output.

*2) Segmentation mask detector*

The segmentation mask detector has class mask layers (12 classes) in 33x33 grids (only Mask R-CNN Inception V2 COCO has 15x15 grid layers) of sigmoid function outputs, namely, the certainty of a corresponding class. The detector adopts the class of "argmax detection class" which is transmitted from the bounding box detector.

As the output shape of those models are either 15x15 or 33x33, the evaluation algorithm extrapolates a mask grid shape to the size of XHWH from the bounding box detector.

*3) Calculation of precisions and recalls by confidence scores*

The detection evaluations (precisions and recalls) are conducted under the decent order of confidence scores at preset maximum detecting instances (300 in this study), and the true/false decision at each instance is made according a threshold (IoU=0.5 in this study). Theoretically if the threshold was increased, a precision increases, and a recall decreases, vise versa.

I. CONCLUSION

There are significant differences in detection performances of bounding boxes and segmentation masks, where the former is constantly superior to the latter.

The holistic evaluation of precisions and recalls by the harmonic average (F1) was made. F1 values of linear cracks, joints, fillings, and shadows are significantly lower in segmentation masks than bounding boxes. Other classes showed similar F1 values.

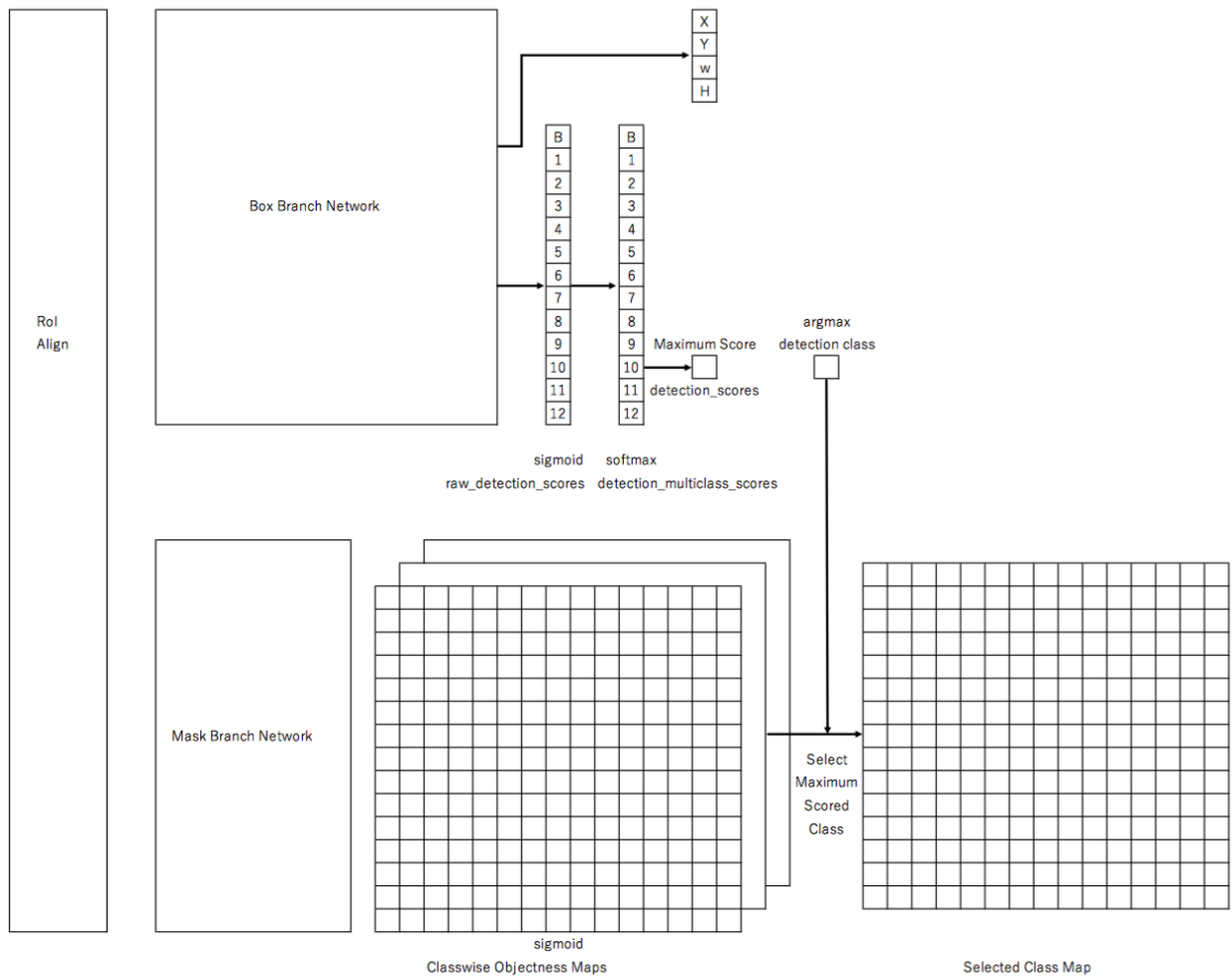

Fig. 9. Schematic illustration of algorithm mechanism outline

Discussions were made on different performances of detection metrics of bounding boxes and segmentation masks focusing the detection algorithms of both detectors.

Typical inferior detection performances of the segmentation mask to the bounding boxes were found, 1) high IoU of bounding boxes vs part of segmentation masks detected, 2) IoU of bounding boxes > 0.5 vs IoU of segmentation masks <0.5, 3) Locations of segmentation masks do not much, and 4) Segmentation masks bigger than ground truths.

Those classes of linear cracks, joints, fillings, and shadows, which are significantly lower detected by segmentation masks are all elongated objects, however, the relationship between detection performances and the shape of the objects is unknown.

ACKNOWLEDGMENT

Special thanks are due to Fukuda Road Co. Ltd for its provision of road images for the annotation works as well as for the provision of information on Multi Fine Eye. We owe much to Tsuyoshi Ikenori for his works on the annotation data. Authors owe to Niigata University of International and Information Studies for funding support on the "Mobile Sensing Development by Deep Learning" project in 2020.